\documentclass[runningheads]{llncs}
\usepackage{graphicx}
\usepackage{algorithmicx,algorithm}
\usepackage{amssymb}
\usepackage{adjustbox}
\usepackage{amsmath}
\usepackage{arydshln}
\usepackage{marvosym}
\usepackage{xcolor}
\newcommand\blfootnote[1]{%
  \begingroup
  \renewcommand\thefootnote{}\footnote{#1}%
  \addtocounter{footnote}{-1}%
  \endgroup
}
\begin{document}
\title{You've Got Two Teachers: Co-evolutionary Image and Report Distillation for Semi-supervised Anatomical Abnormality Detection in Chest X-ray}
%
\titlerunning{Co-evolutionary Semi-supervised Abnormality Detection in Chest X-ray}
%
%
\author{Jinghan Sun\inst{1,2} \and
Dong Wei\inst{2} \and
Zhe Xu \inst{2,3}\and
Donghuan Lu \inst{2}\and
Hong Liu\inst{1,2} \and
Liansheng Wang\inst{1}\textsuperscript{(\Letter)} \and
Yefeng Zheng\inst{2}}
%
%
\authorrunning{J. Sun et al.}
%
%
\institute{National Institute for Data Science in Health and Medicine, Xiamen University, Xiamen, China\\
\email{\{jhsun,liuhong\}@stu.xmu.edu.cn}, \email{lswang@xmu.edu.cn} \and
Tencent Healthcare (Shenzhen) Co., LTD, Tencent Jarvis Lab, Shenzhen, China\\
\email{\{donwei,caleblu,yefengzheng\}@tencent.com, }\and
The Chinese University of Hong Kong, Hong Kong, China\\
\email{jackxz@link.cuhk.edu.hk}
}
%
%
\maketitle              
\begin{abstract}
Chest X-ray (CXR) anatomical abnormality detection aims at localizing and characterising cardiopulmonary radiological findings in the radiographs, which can expedite clinical workflow
and reduce observational oversights\blfootnote{\textsuperscript{*} J. Sun and D. Wei---Contributed equally; J. Sun contributed to this work during an internship at Tencent.}.
Most existing methods attempted this task in either fully supervised settings which demanded costly mass per-abnormality annotations, or weakly supervised settings which still lagged badly behind fully supervised methods in performance. 
In this work, we propose a co-evolutionary image and report distillation (CEIRD) framework, which approaches semi-supervised abnormality detection in CXR by 
grounding the visual detection results with text-classified abnormalities from paired radiology reports, and vice versa.
Concretely, based on the classical teacher-student pseudo label distillation (TSD) paradigm, we additionally introduce an auxiliary report classification model, whose prediction is used 
for report-guided pseudo detection label refinement (RPDLR) in the primary vision detection task.
Inversely, we also use the prediction of the vision detection model for abnormality-guided pseudo classification label refinement (APCLR) in the auxiliary report classification task, and propose a co-evolution strategy where the vision and report models mutually promote each other with RPDLR and APCLR performed alternatively.
To this end, we effectively incorporate the weak supervision by reports into the semi-supervised TSD pipeline.
Besides the cross-modal pseudo label refinement, we further propose an intra-image-modal self-adaptive non-maximum suppression, where the pseudo detection labels generated by the teacher vision model are dynamically rectified by high-confidence predictions by the student.
Experimental results on the public MIMIC-CXR benchmark demonstrate CEIRD's superior performance to several up-to-date weakly and semi-supervised methods.

\end{abstract}

\section{Introduction}
Chest X-ray (CXR) is the most commonly performed diagnostic radiograph in medicine, which helps spot abnormalities or diseases of the airways, blood vessels, bones, heart, and lungs.
Given the complexity and workload of clinical CXR reading, there is a growing interest in developing automated methods for anatomical abnormality detection in CXR \cite{qin2018computer}---especially using deep neural networks (DNNs) \cite{lakhani2017deep,ougul2015lung,rajpurkar2017chexnet}, which are expected to expedite clinical workflow and reduce observational oversights.
Here, the detection task involves both localization (\textit{e.g.}, with bounding boxes) and characterization (\textit{e.g.}, cardiomegaly) of the abnormalities.
However, training accurate DNN-based detection models usually requires large-scale datasets with high-quality per-abnormality annotations, which is costly in time, effort, and expense.

To completely relieve the burden of annotation, a few works \cite{bhalodia2021improving,tam2020weakly,yu2022anatomy} resorted to the radiology reports as a form of weak supervision for localization of pneumonia and pneumothorax in CXR.
The text report describes important findings in each CXR and is available for most archive radiographs, thus is a valuable source of image-level supervision signal unique to medical image data.
However, studies have shown that there are still apparent gaps in performance between image-level weakly supervised and bounding-box-level fully supervised detection \cite{bearman2016s,ji2022point}.
Alternatively, seeking for a trade-off between annotation effort and model performance, semi-supervised learning aims to achieve reasonable performance with an acceptable quantity of manual annotations.
Semi-supervised object detection methods have achieved noteworthy advances in the natural image domain \cite{chen2022label,sohn2020simple,xu2021end}.
Most of these methods 
were built on the teacher-student distillation (TSD) paradigm \cite{hinton2015distilling}, where a teacher model is firstly trained on the labeled data, and then a student model is trained on both the labeled data with real annotations and the unlabeled data with pseudo labels generated (predicted) by the teacher.
However, compared with objects in natural images, the abnormalities in CXR can be subtle and less well-defined with ambiguous boundaries, thus likely to introduce great noise to the pseudo labels and eventually lead to suboptimal performance of semi-supervised learning with TSD.


In this paper, we present a co-evolutionary image and report distillation (CEIRD) framework for semi-supervised anatomical abnormality detection in CXR, incorporating the weak supervision by radiology reports.
Above all, on the basis of TSD \cite{hinton2015distilling}, CEIRD introduces an auxiliary, also semi-supervised, multi-label report classification natural language processing task, whose prediction is used for noise reduction in the pseudo labels of the primary vision detection task, \textit{i.e.}, report-guided pseudo detection label refinement (RPDLR).
Then, noting that the performance of the auxiliary language task is crucial to RPDLR, we inversely use the abnormalities detected in the vision task to filter the pseudo labels in the language task, for abnormality-guided pseudo classification label refinement (APCLR).
In addition, we implement an iterative co-evolution strategy where RPDLR and APCLR are performed alternatively in a loop, where either model is trained while fixing the other and using the other's prediction for pseudo label refinement.
To the best of our knowledge, this is the first work that approaches semi-supervised abnormality detection in CXR by grounding report-classified abnormalities with the visual detection results in the paired radiograph \cite{datta2019align2ground,yang2019fast}, and vice versa.
Besides the cross-modal pseudo label refinement, we additionally propose self-adaptive non-maximum suppression (SA-NMS) for intra-(image-)modal refinement, too, where the predictions by both the teacher and student vision models go through NMS together to produce new pseudo detection labels for training.
In this way, the pseudo labels generated by the teacher are dynamically rectified by high-confidence predictions of the student who is getting better as training goes.
Experimental results on the MIMIC-CXR \cite{johnsonmimic,johnson2019mimic} public benchmark show that our CEIRD outperforms various up-to-date weakly and semi-supervised methods, 
and that its building elements are effective.

To summarize, our contributions include: (1) the complementary RPDLR and APCLR for noise reduction in both vision and language pseudo labels for improved semi-supervised training via mutual grounding, (2) the co-evolution strategy for joint optimization of the primary and auxiliary tasks, and (3) the SA-NMS for dynamic intra-image-modal pseudo label refinement, all contributing to the superior performance of the proposed CEIRD framework.



\section{Method}
\noindent\textbf{Problem Setting.}
In semi-supervised anatomical abnormality localization, a data set comprising both unlabeled samples $D_u=\{(x_i^u,r_i^u)\}_{i=1}^{N_u}$ and labeled ones $D_l=\{(x_i^l,r_i^l,A_i)\}_{i=1}^{N_l}$ is provided for training, where $x$ and $r$ are a CXR and accompanying report, respectively, $A_i = \{(y^l, B^l)\}$ is the annotation for a labeled sample including both bounding boxes $\{B^l\}$ and corresponding categories $\{y^l\}$, and $N^l\ll N^u$ for practical use scenario.
It is worth noting that $\{y^l\}$ can also be considered as  classification labels for the report.
The objective is to obtain a detection model that can accurately localize and classify the abnormalities in any testing CXR (without report in practice), by making good use of both the labeled and unlabeled CXRs plus the accompanying reports in the training set.

\begin{figure}[t]
\centering
\includegraphics[width=.95\textwidth,height=5cm]{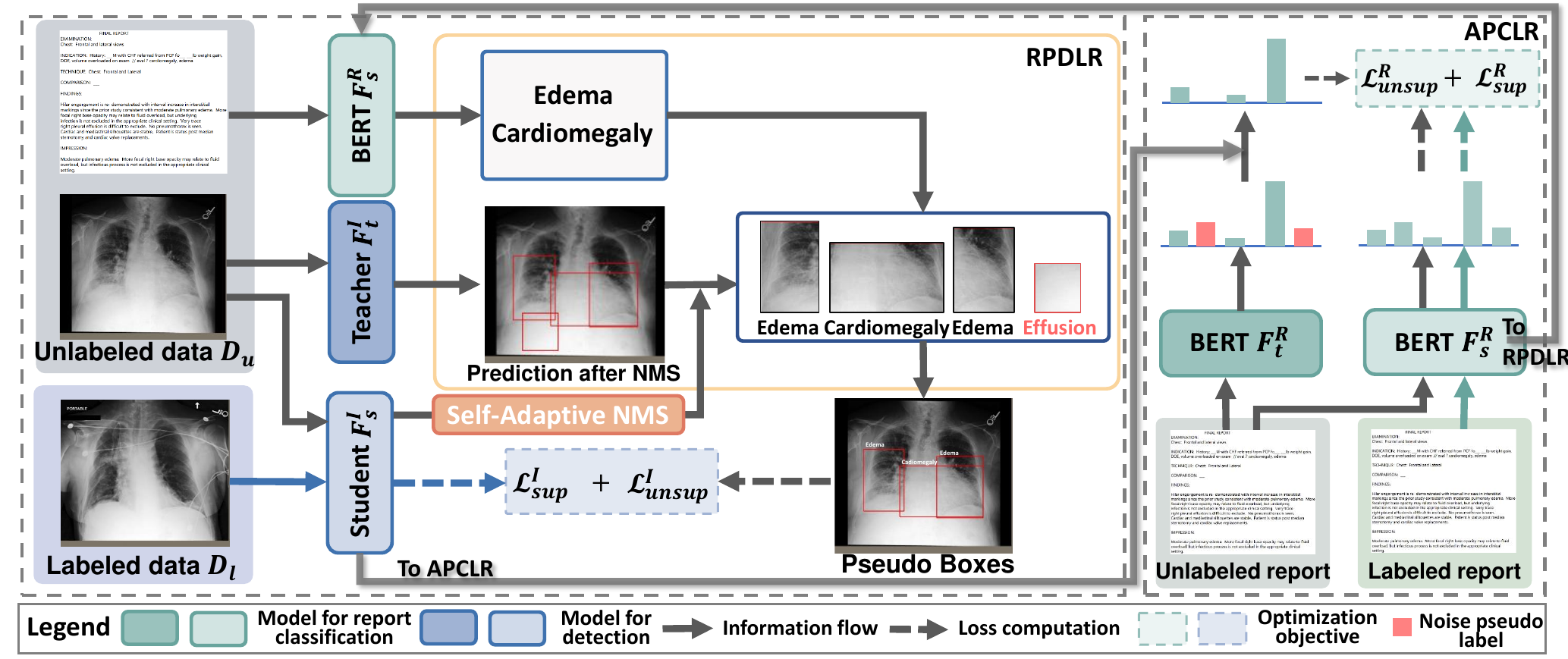}
\caption{Overview of the proposed framework.
RPDLR: report-guided pseudo detection label refinement;
APCLR: abnormality-guided pseudo classification label refinement.}
\label{fig:procedure}
\end{figure}

\vspace{3.5mm}\noindent\textbf{Method Overview.}
Fig.~\ref{fig:procedure} provides an overview of our framework. 
Suppose a pretrained teacher vision model $F^I_t$ (\textit{e.g.}, on labeled data) 
for abnormality detection in CXR is given, together with a pretrained language model $F_s^R$ 
for multi-label abnormality classification of reports.
On the one hand, we generate for an unlabeled image $x_i^u$ pseudo detection labels with $F^I_t$ and filter the pseudo labels by self-adaptive non-maximum suppression (NMS).
Meanwhile, we feed the corresponding report $r_i^u$ into $F^R_s$ and use the   prediction for report-guided pseudo detection label refinement (RPDLR).
To this end, we obtain refined pseudo labels to supervise the student vision model $F_s^I$ toward better anatomical abnormality localization.
On the other hand, we also pass the detection predictions by $F_s^I$ to a teacher language model $F_t^R$ for abnormality-guided pseudo classification label refinement (APCLR), to better supervise the student language model $F_s^R$ on unlabeled data for report-based abnormality classification.
In turn, the better language model $F_s^R$ helps train better vision models via RPDLR,
thus both types of models co-evolve during training.
Note that the real labels are used to train both student models along with the pseudo ones.
After training, we only need the student vision model $F_s^I$ for abnormality localization in testing CXRs.

\vspace{3.5mm}\noindent\textbf{Preliminary Pseudo Label Distillation for Semi-supervised Learning.}\label{sec:method:preliminary}
Both of our baseline semi-supervised vision and language models follow the teacher-student knowledge distillation (TSD) procedure \cite{hinton2015distilling}.
For report classification, we first train a teacher model $F_t^R$ on labeled reports, and then train a student model $F_s^R$ to predict real labels on labeled reports and pseudo labels produced by $F_t^R$ on unlabeled ones with the loss function:
\begin{equation}\label{eq:semi-class}
    \mathcal{L}^{R} = \mathcal{L}_\mathrm{sup}^{R} + \mathcal{L}_\mathrm{unsup}^{R} =  {\sum}_{D_l}\mathcal{L}_\mathrm{cls}^{R}\left({\hat{y}^l}, {y^l}\right) + {\sum}_{D_u}\mathcal{L}_\mathrm{cls}^{R}\left({\hat{y}}, {y^{pr}}\right),
\end{equation}
where $\mathcal{L}_\mathrm{cls}^{R}$ is the cross-entropy loss, $\{\hat{y}\}=F^R_s(r)$ is the prediction by the student model, $\{y^{pr}\}=F_t^R(r^u)$ is the set of pseudo labels generated by the teacher model.
In each batch, labeled and unlabeled instances are sampled according to a controlled ratio.
The resulting report classification model $F_s^R$ will be utilized later to help with the primary task of abnormality detection in CXR.
Similarly, a student vision model $F_s^I$ for abnormality detection in CXR is trained in semi-supervised setting by distilling from a teacher vision model $F_t^I$ trained on labeled CXRs, with the loss function:
\begin{equation}
\label{eq:pld}
\begin{aligned}
    \mathcal{L}^{I} = \mathcal{L}_\mathrm{sup}^{I} + \mathcal{L}_\mathrm{unsup}^{I}
    ={\sum}_{D_l}\big[\mathcal{L}_\mathrm{cls}^{I}\left({\hat{y}^l}, {y^l}\right)+ \mathcal{L}_\mathrm{reg}^{I}\big({\hat{B}^l}, {B^l}\big)\big]\\
+{\sum}_{D_u}\big[\mathcal{L}_\mathrm{cls}^{I}\left({\hat{y}}, {y^{pv}}\right)+ \mathcal{L}_\mathrm{reg}^{I}\big({\hat{B}}, {B^{pv}}\big)\big],
\end{aligned}
\end{equation}
where $\{(\hat{y},\hat{B})\}=F^I_s(x)$ are the predictions by the student model,
$\{(y^{pv},B^{pv})\}=F_t^I(x^u)$ are the pseudo class and bounding box labels generated by the teacher model, $\mathcal{L}_\mathrm{cls}^{I}$ is the focal loss \cite{lin2017focal} for abnormality classification, and $\mathcal{L}_\mathrm{reg}^{I}$ is the smooth L1 loss for bounding box regression.

\vspace{1.75mm}\noindent\textbf{Self-Adaptive Non-maximum Suppression.}
During the TSD, the teacher vision model $F_t^I$ is 
kept fixed.
While its knowledge suffices for guiding the student vision model $F_s^I$ in the early stage of TSD, it may somehow impede the learning of $F_s^I$ when $F_s^I$ gradually improves by also learning from the large amount of unlabeled data.
Therefore, to gradually improve quality and robustness of the pseudo detection labels as $F_s^I$ learns,
we propose to perform self-adaptive non-maximum suppression (SA-NMS) to combine the pseudo labels $\{(y^{pv},B^{pv})\}$ output by $F_t^I$ and the predictions $\{(\hat{y},\hat{B})\}$ by $F_s^I$ in each mini batch.
Specifically, we perform NMS on the combined set of the pseudo labels and predictions: $\{(y^{cv},B^{cv})\}= \mathrm{NMS}\big(\{(y^{pv},B^{pv})\} \bigcup \{(\hat{y}, \hat{B})\}\big)$, and replace $\{(y^{pv},B^{pv})\}$ in Eq. (\ref{eq:pld}) with $\{(y^{cv},B^{cv})\}$ for supervision by unlabeled CXRs.
In this way, highly confident predictions by the maturing student can rectify imprecise ones by the teacher, leading to better supervision signals stemming from unlabeled data.

\vspace{1.5mm}\noindent\textbf{Report-guided Pseudo Label Refinement.}
In routine clinics, almost every radiograph in archive is accompanied by a report describing findings, abnormalities (if any), and diagnosis.
Compared with the captions of natural images, the report texts constitute a unique (to medical image analysis) and rich source of extra information in addition to the image modality.
To this end, we propose report-guided pseudo detection label refinement (RPDLR) to make use of this cross-modal information for semi-supervised anatomical abnormality detection in CXR.
Specifically, we use the student language
model $F_s^R$ (trained with Eq. (\ref{eq:semi-class})) to refine the pseudo detection labels.
Given a pair of unlabeled image $x^u$ and report $r^u$, we obtain the set of abnormalities $\{({y^{cv}},{B^{cv}})\}$ detected in $x^u$ after SA-NMS, and the set of abnormalities $\{\hat{y}\}$ classified in $r^u$ by $F_s^R$.
Then, we only keep the pseudo detection labels whose categories are in the report-classified abnormalities:
\begin{equation}\label{eq.filter-det}
  \{(y^v,B^v)\} = \left. \left\{\left({y^{cv}_j},{B^{cv}_j}\right) \right| {y^{cv}_j} \in \{\hat{y}\}\right\}.
\end{equation}
Eventually, we train the student vision model $F_s^I$ using $\{(y^v,B^v)\}$ in Eq. (\ref{eq:pld}).

\vspace{2.5mm}\noindent\textbf{Co-evolutionary Semi-supervised Learning with Cycle Pseudo Label Refinement.}
As the auxiliary student language model $F_s^R$ plays an important role in RPDLR, it is reasonable to optimize its performance which in turn would benefit the primary task of abnormality detection.
Therefore, we further propose an inverse, abnormality-guided pseudo classification labels refinement (APCLR) to help with semi-supervised training of the report classification model.
Similarly in concept to the RPDLP, given a pair of unlabeled image $x^u$ and report $r^u$, we obtain the set of abnormalities $\{(\hat{y},\hat{B})\}$ detected in $x^u$ by the student vision model $F_s^I$, and the set of classification pseudo labels $\{y^{pr}\}$ generated for $r^u$ by the teacher language model $F_t^R$.
We retain only the pseudo labels $\{y^{pr}_j|y^{pr}_j\in\{\hat{y}\}\}$, by excluding the report-classified abnormalities not detected in the paired CXR.

\begin{figure}[t]
\centering
\includegraphics[width=.95\textwidth]{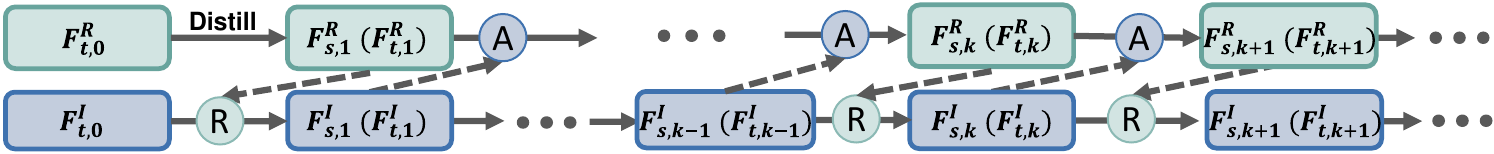}
\caption{Illustration of the co-evolution strategy.
``R'' and ``A'' represent report- and abnormality-guided pseudo label refinements (RPDLR and APCLR), respectively.
}
\label{fig:co-evolution}
\end{figure}

Ideally, one should use an optimal report classification model for refinement of the abnormality detection pseudo labels, and vice versa.
However, the two models are mutually dependent on each other in a circle.
To solve this dilemma, we implement an alternative co-evolution strategy to refine the abnormality detection and report classification pseudo labels iteratively, in \textit{generations}.
As shown in Fig. \ref{fig:co-evolution}, the $k$\textsuperscript{th} generation student vision model $F_{s,k}^I$ is distilled from the teacher $F_{t,k-1}^I$, whose pseudo labels are refined by the prediction of the frozen student language model $F_{s,k}^R$ via RPDLR.
Subsequently, $F_{s,k}^I$ is frozen and used to 1) help train the $(k+1)$\textsuperscript{th} student language model $F_{s,k+1}^R$ via APCLR, and 2) serve as the teacher vision model in next generation: $F_{s,k}^I\rightarrow F_{t, k}^I$.\footnote{The initial teachers $F_{t,0}^I$ and $F_{t,0}^R$ are obtained by training on the labeled data only.}
Note that in each generation the students are reborn with random initialization \cite{furlanello2018born}. 
Thus the co-evolution continues to optimize the vision and report models cyclically with cross-modal mutual promotion.
After training, we only need the $K$\textsuperscript{th} generation student vision model $F_{s,K}^I$ for abnormality detection in upcoming test CXRs.

\section{Experiments}
\noindent\textbf{Dataset and Evaluation Metrics.}
We conduct experiments on the chest radiography dataset MIMIC-CXR \cite{johnsonmimic,johnson2019mimic}, with the detection annotations provided by MS-CXR \cite{boecking2022making}.
MIMIC-CXR is a large publicly available dataset of CXR and free-text radiology reports. 
MS-CXR provides bounding box annotations for part of the CXRs in MIMIC-CXR (1,026 samples).
MIMIC-CXR includes 14 categories of anatomical abnormalities for multi-label classification of the reports, while there are only eight categories in the bounding box annotations of MS-CXR.
For consistency, we exclude samples in MIMIC-CXR that have abnormality labels outside the eight categories of MS-CXR, leaving 112,425 samples.
Thus, in our semi-supervised setting, 1,026 samples are labeled and the rest are not.\footnote{In this work, we deliberately construct the semi-supervised setting by ignoring the report labels provided in MIMIC-CXR for methodology development.
When it comes to a practical new application with no such label available, \textit{e.g.}, semi-supervised lesion detection in color fundus photography, our method can be readily applied.} 
We split the labeled samples for training, validation, and testing according to the ratio of 7:1:2, and use the remaining samples as our unlabeled training data.
We focus on the frontal views in this work.
The mean average precision (mAP) \cite{everingham2009pascal} with the intersection over union (IoU) threshold of 0.25, 0.5, and 0.75 is employed to evaluate the performance of abnormality detection in CXR.

\vspace{1.5mm}\noindent\textbf{Implementation.}
The PyTorch \cite{paszke2019pytorch} framework (1.4.0) is used for experiments. For report classification, we employ the BERT-base uncased model \cite{devlin2018bert} with eight linear heads.
Stochastic gradient descent with the momentum of 0.9 and learning rate of $10^{-4}$ is used for optimization.
The batch size is set to 16 reports. 
For abnormality detection, we employ RetinaNet \cite{lin2017focal} with FPN \cite{lin2017feature}+ResNet-101 \cite{he2016deep} as backbone. We resize all images to 512$\times$512 pixels and use a batch size
of 16. 
Data augmentation including random cropping and flipping is performed.
Our implementation and hyper-parameters follow the official settings \cite{lin2017focal}.
Unless otherwise stated, we evolve the vision and language models for {two} generations, and train both models for 2000 iterations in each generation (including initial training of the teacher models).
{The ratio of labeled to unlabeled samples in each mini batch during the semi-supervised training is empirically set to 1:1 and 2:1 for the language and vision models, respectively.}
The source code is available at: https://github.com/jinghanSunn/CEIRD.



\begin{table}[t]
\caption{Abnormality detection results on the test data, using mAP (\%) with the IoU thresholds of 0.25, 0.5, and 0.75.
TSD: teacher-student distillation.}
\begin{adjustbox}{width=\textwidth}
\begin{tabular}{l|cc;{2pt/3pt}c;{2pt/3pt}ccccc;{2pt/3pt}c}
\hline
{mAP}           & {CAM \cite{zhou2016learning}} & {AGXNet \cite{yu2022anatomy}} & {Sup.} & {TSD \cite{hinton2015distilling}} & {STAC \cite{sohn2020simple}} & {LabelMatch \cite{chen2022label}} & {Soft Teacher \cite{xu2021end}} & {Ours}     & {Semi-oracle} \\ \hline
{@0.25} & {20.47}                      & {29.96}                      & {37.91}     & {38.29}                               & {39.26}                    & {39.92}                         & {40.17}                       & {\textbf{41.93}} & 42.61                           \\ 
{@0.5}  & {11.20}                      & {15.62}                      & {32.84}     & {33.95}                               & {35.01}                    & {36.40}                         & {36.59}                       & {\textbf{37.20}} & 37.39                           \\ 

{@0.75} & {3.05}                       & {7.44}                       & {19.21}     & {19.51}                               & {23.90}                    & {24.06}                         & {24.78}                       & {\textbf{25.12}} & 25.66                           \\ \hline
\end{tabular}
\end{adjustbox}
\label{tab:det_res}
\end{table}

\vspace{1.5mm}\noindent\textbf{Comparison with State-of-the-Art (SOTA) Methods.}
We compare our proposed co-evolution image and report distillation (CEIRD) framework
with several up-to-date detection methods, including weakly supervised: CAM \cite{zhou2016learning} (locating objects based on class activation maps), AGXNet \cite{yu2022anatomy} (aiding CAM-based localization with report representations), fully supervised on labeled training data only (Sup.), baseline semi-supervised via teacher-student pseudo-label distillation  (TSD; see Eq. (\ref{eq:pld})) \cite{hinton2015distilling}, and three SOTA semi-supervised (STAC \cite{sohn2020simple}, LabelMatch \cite{chen2022label}, and Soft Teacher \cite{xu2021end}) object detection methods.

The results are shown in Table~\ref{tab:det_res}, from which we have the following observations. 
First, both fully supervised (by the labeled data only) and semi-supervised methods outperform the weakly-supervised by large margins,
proving the efficacy of using limited annotations.
Second, all semi-supervised methods outperform the fully supervised (by the labeled data only) by various margins, demonstrating apparent benefit of making use of the unlabeled data, too.
Third, our CEIRD achieves the best performance of all the semi-supervised methods for the mAPs evaluated at three different IoU thresholds, outperforming the second best (Soft Teacher) by {up to 1.76\%}.
These results clearly demonstrate the advantage of our method which innovatively integrates the semi-supervision by unlabeled images and the weak supervision by texts.
{In addition, we evaluate a semi-oracle of our method, where the ground truth report labels provided in MIMIC-CXR are used for RPDLR, instead of the auxiliary model's prediction.
As we can see, our method is marginally short of the semi-oracle, \textit{e.g.}, {37.20}\% versus 37.39\% for mAP@0.5, suggesting that our co-evolution strategy can effectively mine the relevant information from the reports.}
We provide in the supplementary material visualizations of example detection results by Soft Teacher and our method, where ours are visually superior with fewer misses (left), fewer false positives (middle), and better localization (right).
{Lastly, we also provide performance evaluation of the auxiliary report classification task in the supplement.}

\begin{figure}[t]
\begin{minipage}[c]{0.36\textwidth} 
\begin{table}[H]
    \centering
    \caption{Ablation study results on the {validation} data.}\label{tab:ablation}
    \begin{adjustbox}{width=1\textwidth}
    \begin{tabular}{c|c:c:c:c}
    \hline
        &      Baseline     & (a)        & (b)                  & (c)            \\ \hline
    RPDLR                   & - & \checkmark & \checkmark & \checkmark     \\
    CoE+APCLR               &  -                     & -  & \checkmark           & \checkmark     \\
    SA-NMS                    & -                      & -           &       -               & \checkmark     \\ \hline
mAP@0.25  & 37.31    & 39.10      & 39.72      & \textbf{41.86} \\
mAP@0.5   & 32.76    & 35.87      & 35.90      & \textbf{36.18} \\
mAP@0.75  & 17.96    & 22.02      & 23.55      & \textbf{24.49} \\ \hline
    \end{tabular}
    \end{adjustbox}
\end{table}
\end{minipage}%
\hfill%
\begin{minipage}[c]{0.58\textwidth}
\begin{figure}[H]
    \centering 
    \includegraphics[width=.95\textwidth,height=2cm]{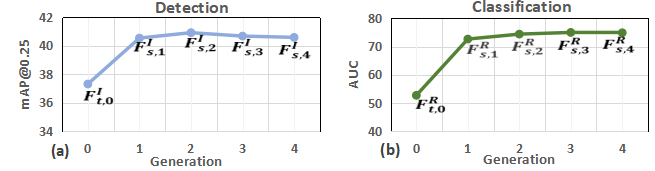} 
    \setlength{\abovecaptionskip}{0.3cm}
    \caption{Performance of (a) vision and (b) report models as a function of generations.
    AUC: area under the receiver operating characteristic curve.}
    \label{fig:generation}
\end{figure}
\end{minipage}
\end{figure}

\vspace{1.5mm}\noindent\textbf{Ablation Study.}
We conduct ablation studies on the {validation data} to investigate efficacy of the novel building elements of our CEIRD framework, including:
report-guided pseudo detection label refinement (RPDLR), 
co-evolution strategy (CoE) with abnormality-guided pseudo classification label refinement (APCLR), and 
self-adaptive non-maximum suppression (SA-NMS).
We use the preliminary teacher-to-student pseudo label distillation as baseline (Eq. (\ref{eq:pld})).
As shown in Table \ref{tab:ablation}, RPDLR (column (a)) substantially boosts performance upon the baseline by {1.79--4.06\%} in mAPs thanks to the refined pseudo detection labels.
By adopting CoE+APCLR (column (b)), we achieve further performance improvements {up to 1.53\%} as the auxiliary report classification model gets better together.
Last but not the least, the introduction of SA-NMS (column (c)) also brings improvements {up to 2.14\%}.
These results validate the novel design of our framework.
In addition, we conduct experiments to empirically determine the optimal number of  generations for the co-evolution.
The results are shown in Fig. \ref{fig:generation}, where the 0\textsuperscript{th} generation means fully supervised models trained on the labeled data only (\textit{i.e.}, the initial teacher models $F_{t,0}^I$ and $F_{t,0}^R$).
{As we can see, 
the vision and report models improve in the first two and three generations, respectively, and then remain stable in the following ones, confirming that
both models promote each other with the co-evolution strategy.
Since our ultimate objective is abnormality detection in CXR, we select two generations for comparison with other methods.}

\section{Conclusion}
In this work, we proposed a new co-evolutionary image and report distillation (CEIRD) framework for semi-supervised anatomical abnormality detection in chest X-ray.
On the basis of a preliminary teacher-student pseudo label distillation, we first presented self-adaptive NMS to mingle highly confident predictions by both the teacher and student for improved pseudo labels.
We then proposed report-guided pseudo detection label refinement (RPDLR) that used abnormalities classified from the accompanying radiology reports by an auxiliary language model to eliminate unmatched pseudo labels.
Meanwhile, 
we further proposed an inverse, abnormality-guided pseudo classification label refinement (APCLR) making use of the abnormalities detected in X-ray images for better language model training.
In addition, we implemented a co-evolution strategy that looped the RPDLR and APCLR to iteratively optimize the main vision detection model and auxiliary report classification model in an alternative manner.
Experimental results showed that our CEIRD framework achieved superior performance to up-to-date semi-/weakly-supervised methods.
\subsubsection{Acknowledgement.} This work was supported by the National Key R\&D Program of China under Grant 2020AAA0109500/2020AAA0109501 and the National Key Research and Development Program of China (2019YFE0113900).
%
%
\bibliographystyle{splncs04}
\bibliography{mybibliography}

\begin{thebibliography}{10}
\providecommand{\url}[1]{\texttt{#1}}
\providecommand{\urlprefix}{URL }
\providecommand{\doi}[1]{https://doi.org/#1}

\bibitem{bearman2016s}
Bearman, A., Russakovsky, O., Ferrari, V., Fei-Fei, L.: What's the point:
  Semantic segmentation with point supervision. In: Proceedings of the European
  Conference on Computer Vision. pp. 549--565. Springer (2016)

\bibitem{bhalodia2021improving}
Bhalodia, R., Hatamizadeh, A., Tam, L., Xu, Z., Wang, X., Turkbey, E., Xu, D.:
  Improving pneumonia localization via cross-attention on medical images and
  reports. In: Proceedings of the International Conference on Medical Image
  Computing and Computer Assisted Intervention, Part II. pp. 571--581. Springer
  (2021)

\bibitem{boecking2022making}
Boecking, B., Usuyama, N., Bannur, S., Castro, D.C., Schwaighofer, A., Hyland,
  S., Wetscherek, M., Naumann, T., Nori, A., Alvarez-Valle, J., et~al.: Making
  the most of text semantics to improve biomedical vision--language processing.
  arxiv Preprint arxiv:2204.09817  (2022)

\bibitem{chen2022label}
Chen, B., Chen, W., Yang, S., Xuan, Y., Song, J., Xie, D., Pu, S., Song, M.,
  Zhuang, Y.: Label matching semi-supervised object detection. In: Proceedings
  of the IEEE/CVF Conference on Computer Vision and Pattern Recognition. pp.
  14381--14390 (2022)

\bibitem{datta2019align2ground}
Datta, S., Sikka, K., Roy, A., Ahuja, K., Parikh, D., Divakaran, A.:
  {Align2Ground}: Weakly supervised phrase grounding guided by image-caption
  alignment. In: Proceedings of the IEEE/CVF International Conference on
  Computer Vision. pp. 2601--2610 (2019)

\bibitem{devlin2018bert}
Devlin, J., Chang, M.W., Lee, K., Toutanova, K.: {BERT}: Pre-training of deep
  bidirectional transformers for language understanding. arxiv Preprint
  arxiv:1810.04805  (2018)

\bibitem{everingham2009pascal}
Everingham, M., Van~Gool, L., Williams, C.K., Winn, J., Zisserman, A.: The
  {PASCAL} visual object classes {(VOC)} challenge. International Journal of
  Computer Vision  \textbf{88},  303--308 (2009)

\bibitem{furlanello2018born}
Furlanello, T., Lipton, Z., Tschannen, M., Itti, L., Anandkumar, A.: Born again
  neural networks. In: International Conference on Machine Learning. pp.
  1607--1616. PMLR (2018)

\bibitem{he2016deep}
He, K., Zhang, X., Ren, S., Sun, J.: Deep residual learning for image
  recognition. In: Proceedings of the IEEE Conference on Computer Vision and
  Pattern Recognition. pp. 770--778 (2016)

\bibitem{hinton2015distilling}
Hinton, G., Vinyals, O., Dean, J.: Distilling the knowledge in a neural
  network. arxiv Preprint arxiv:1503.02531  (2015)

\bibitem{ji2022point}
Ji, H., Liu, H., Li, Y., Xie, J., He, N., Huang, Y., Wei, D., Chen, X., Shen,
  L., Zheng, Y.: Point beyond class: A benchmark for weakly semi-supervised
  abnormality localization in chest {X}-rays. In: Proceedings of the
  International Conference on Medical Image Computing and Computer Assisted
  Intervention, Part III. pp. 249--260. Springer (2022)

\bibitem{johnsonmimic}
Johnson, A., Lungren, M., Peng, Y., Lu, Z., Mark, R., Berkowitz, S., Horng, S.:
  {MIMIC-CXR-JPG}-chest radiographs with structured labels. PhysioNet  (2019)

\bibitem{johnson2019mimic}
Johnson, A.E., Pollard, T.J., Greenbaum, N.R., Lungren, M.P., Deng, C.y., Peng,
  Y., Lu, Z., Mark, R.G., Berkowitz, S.J., Horng, S.: {MIMIC-CXR-JPG}, a large
  publicly available database of labeled chest radiographs. arxiv Preprint
  arxiv:1901.07042  (2019)

\bibitem{lakhani2017deep}
Lakhani, P., Sundaram, B.: Deep learning at chest radiography: automated
  classification of pulmonary tuberculosis by using convolutional neural
  networks. Radiology  \textbf{284}(2),  574--582 (2017)

\bibitem{lin2017feature}
Lin, T.Y., Doll{\'a}r, P., Girshick, R., He, K., Hariharan, B., Belongie, S.:
  Feature pyramid networks for object detection. In: Proceedings of the IEEE
  Conference on Computer Vision and Pattern Recognition. pp. 2117--2125 (2017)

\bibitem{lin2017focal}
Lin, T.Y., Goyal, P., Girshick, R., He, K., Doll{\'a}r, P.: Focal loss for
  dense object detection. In: Proceedings of the IEEE International Conference
  on Computer Vision. pp. 2980--2988 (2017)

\bibitem{ougul2015lung}
O{\u{g}}ul, B.B., Ko{\c{s}}ucu, P., {\"O}z{\c{c}}am, A., Kanik, S.D.: Lung
  nodule detection in {X}-ray images: a new feature set. In: 6th European
  Conference of the International Federation for Medical and Biological
  Engineering: MBEC 2014, 7-11 September 2014, Dubrovnik, Croatia. pp.
  150--155. Springer (2015)

\bibitem{paszke2019pytorch}
Paszke, A., Gross, S., Massa, F., Lerer, A., Bradbury, J., Chanan, G., Killeen,
  T., Lin, Z., Gimelshein, N., Antiga, L., et~al.: {PyTorch}: An imperative
  style, high-performance deep learning library. Advances in Neural Information
  Processing Systems  \textbf{32} (2019)

\bibitem{qin2018computer}
Qin, C., Yao, D., Shi, Y., Song, Z.: Computer-aided detection in chest
  radiography based on artificial intelligence: a survey. Biomedical
  Engineering Online  \textbf{17}(1),  1--23 (2018)

\bibitem{rajpurkar2017chexnet}
Rajpurkar, P., Irvin, J., Zhu, K., Yang, B., Mehta, H., Duan, T., Ding, D.,
  Bagul, A., Langlotz, C., Shpanskaya, K., et~al.: {CheXNet}: Radiologist-level
  pneumonia detection on chest {X}-rays with deep learning. arxiv preprint
  arxiv:1711.05225  (2017)

\bibitem{sohn2020simple}
Sohn, K., Zhang, Z., Li, C.L., Zhang, H., Lee, C.Y., Pfister, T.: A simple
  semi-supervised learning framework for object detection. arxiv Preprint
  arxiv:2005.04757  (2020)

\bibitem{tam2020weakly}
Tam, L.K., Wang, X., Turkbey, E., Lu, K., Wen, Y., Xu, D.: Weakly supervised
  one-stage vision and language disease detection using large scale pneumonia
  and pneumothorax studies. In: Proceedings of the International Conference on
  Medical Image Computing and Computer Assisted Intervention, Part IV 23. pp.
  45--55. Springer (2020)

\bibitem{xu2021end}
Xu, M., Zhang, Z., Hu, H., Wang, J., Wang, L., Wei, F., Bai, X., Liu, Z.:
  End-to-end semi-supervised object detection with soft teacher. In:
  Proceedings of the IEEE/CVF International Conference on Computer Vision. pp.
  3060--3069 (2021)

\bibitem{yang2019fast}
Yang, Z., Gong, B., Wang, L., Huang, W., Yu, D., Luo, J.: A fast and accurate
  one-stage approach to visual grounding. In: Proceedings of the IEEE/CVF
  International Conference on Computer Vision. pp. 4683--4693 (2019)

\bibitem{yu2022anatomy}
Yu, K., Ghosh, S., Liu, Z., Deible, C., Batmanghelich, K.: Anatomy-guided
  weakly-supervised abnormality localization in chest {X}-rays. In: Proceedings
  of the International Conference on Medical Image Computing and Computer
  Assisted Intervention, Part V. pp. 658--668. Springer (2022)

\bibitem{zhou2016learning}
Zhou, B., Khosla, A., Lapedriza, A., Oliva, A., Torralba, A.: Learning deep
  features for discriminative localization. In: Proceedings of the IEEE
  Conference on Computer Vision and Pattern Recognition. pp. 2921--2929 (2016)

\end{thebibliography}

\end{document}